\documentclass[runningheads]{llncs}
\usepackage{graphicx}
\usepackage[colorlinks,citecolor=blue,urlcolor=blue]{hyperref}

\usepackage{amsmath}
\usepackage{amssymb}

\def\argmax{\operatornamewithlimits{arg\,max}}

\usepackage{multirow}
\usepackage{booktabs}
\usepackage{colortbl}
\usepackage{xcolor}
\usepackage{adjustbox}
\usepackage{cleveref}

\usepackage{xspace}
\makeatletter
\DeclareRobustCommand\onedot{\futurelet\@let@token\@onedot}
\def\@onedot{\ifx\@let@token.\else.\null\fi\xspace}
\def\eg{\emph{e.g}\onedot} 
\def\ie{\emph{i.e}\onedot} 
 
\def\etc{\emph{etc}\onedot}

\makeatother

\newcommand{\rf}[1]{\textbf{\textcolor{red}{#1}}}
\newcommand{\rs}[1]{\textbf{\textcolor{green}{#1}}}
\newcommand{\rt}[1]{\textbf{\textcolor{blue}{#1}}}
\newcommand{\ours}{\cellcolor{red!20}}
\newcommand{\lniqe}[1]{\textbf{#1}}

\begin{document}
\title{AdvHaze: Adversarial Haze Attack}
%
%
\author{
Ruijun Gao\inst{1}\orcidID{0000-0002-0654-2455} \and
Qing Guo\inst{2}\orcidID{0000-0003-0974-9299} \and \\
Felix Juefei-Xu\inst{3}\orcidID{0000-0002-0857-8611} \and
Hongkai Yu\inst{4}\orcidID{0000-0001-5383-8913} \and
Wei Feng\inst{1}\orcidID{0000-0003-3809-1086}
}

\authorrunning{R. Gao et al.}

\institute{
College of Intelligence and Computing, Tianjin University, Tianjin 300350, China\\
\email{gaoruijun@tju.edu.cn,wfeng@ieee.org}\\
\and
Nanyang Technological University, Singapore 639798, Singapore\\
\email{qing.guo@ntu.edu.sg}
\and
Alibaba Group, USA\\
\and
Cleveland State University, Cleveland 44115, USA\\
\email{h.yu19@csuohio.edu}
}

\maketitle

\begin{abstract}
In recent years, adversarial attacks have drawn more attention for their value on evaluating and improving the robustness of machine learning models, especially, neural network models.
However, previous attack methods have mainly focused on applying some $l^p$ norm-bounded noise perturbations. In this paper, we instead introduce a novel adversarial attack method based on haze, which is a common phenomenon in real-world scenery. Our method can synthesize potentially adversarial haze into an image based on the atmospheric scattering model with high realisticity and mislead classifiers to predict an incorrect class.
We launch experiments on two popular datasets, \ie, ImageNet and NIPS~2017. We demonstrate that the proposed method achieves a high success rate, and holds better transferability across different classification models than the baselines. We also visualize the correlation matrices, which inspire us to jointly apply different perturbations to improve the success rate of the attack.
We hope this work can boost the development of non-noise-based adversarial attacks and help evaluate and improve the robustness of DNNs.

\end{abstract}

\section{Introduction}

\begin{figure}[t]
\centering
\includegraphics[width=0.9\textwidth]{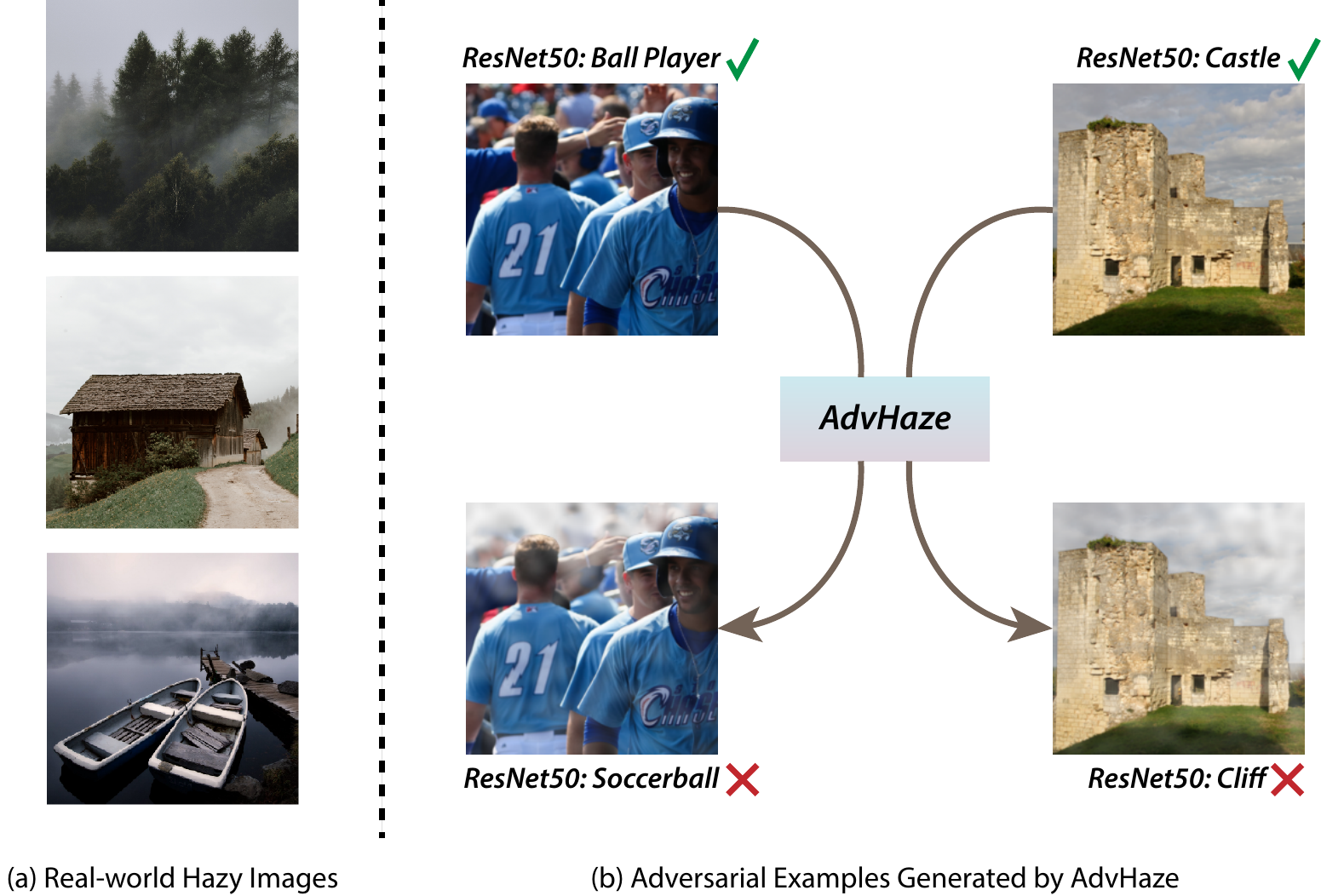}
\caption{(a) shows three hazy images captured in real-world scenes. (b) shows the adversarial examples generated by AdvHaze mislead a ResNet-50 \cite{resnet} to predict an incorrect class, thanks to the synthesized hazy phenomenon.}\label{fig:overall}
\end{figure}

Deep neural networks (DNN) have become extremely successful in many computer vision tasks, \eg, image recognition, co-saliency detection, visual object tracking, \etc. However, the existing DNNs have also shown their vulnerability to adversarial examples, which are carefully crafted by adding imperceptible noises or applying natural transformations. Adversarial examples help evaluate and improve the robustness of various machine learning models.

Haze is a common natural phenomenon led by the complex atmospheric conditions and light sources, widely existing in wildly captured images. Haze synthesis is an important task in the field of computer vision. However, there is no related work, which leverages haze synthesis to accomplish an adversarial attack against DNNs. To this end, we propose an adversarial haze attack, as shown in Figure~\ref{fig:overall}. Our method can effectively synthesize haze into an image and mislead classifiers to predict an incorrect class. We hope this can bridge the gap and boost research on novel adversarial attacks. The major contributions can be summarized as follows:

\begin{itemize}
  \item We propose a novel adversarial attack method, \ie, AdvHaze. This attack generates adversarial examples by synthesizing potentially adversarial haze with high realisticity, to mislead classifiers to predict an incorrect class.
  \item We conduct comprehensive experiments to evaluate the performance of the proposed method and baseline attacks and image quality of the adversarial examples. This has demonstrated the effectiveness of the proposed method.
  \item We visualize and discuss the correlation among attacks in our experimental setup and draw conclusions in agreement with our intuition.
\end{itemize}

\subsubsection{Related Work.} In recent several years, adversarial attacks have been proposed to fool a DNN by adding imperceptible noises or applying natural transformations to an image input. Among them, several classical adversarial noise attacks have shown promising results, \eg, gradient based fast gradient sign method (FGSM) \cite{fgsm}, basic iteration method (BIM, also known as iterative FGSM) \cite{ifgsm}, distance metric based C\&W method \cite{cw}, momentum iterative fast gradient sign method (MI-FGSM) \cite{mifgsm}, and \etc. Some studies tried to apply natural transformation to an image for attack, which is imperceptible to humans for real-world applications as well, \eg, semantically aware colorization based cAdv and texture transfer based tAdv \cite{cadv}, adversarial watermark based AdvWatermark \cite{advwatermark}, motion blurring based ABBA \cite{abba}, adversarial rain attack based AdvRain \cite{advrain}, \etc. The AdvHaze method proposed in this paper belongs to a novel natural transformation based method by adding the adversarial haze for attack.   



\section{Methodology}

\subsection{Haze Synthesis}

In the field of computer graphics and computer vision, the atmospheric scattering model is widely used to synthesize a hazy image, \ie,
\begin{align}\label{eq:haze}
    \mathbf{H}(\mathbf{x}) = \mathbf{I}(\mathbf{x})t(\mathbf{x}) + \mathbf{A}(\mathbf{x})(1 - t(\mathbf{x})),
\end{align}
where $\mathbf{x}$ represents any location in an image, $\mathbf{t}$ is the medium transmission rate describing the portion of the light that reaches the camera, $\mathbf{A}$ is the global atmospheric light, $\mathbf{I}$ is the scene radiance, \ie, the haze-free image, and $\mathbf{H}$ is the synthesized hazy image, which is captured by the camera. Especially, given the scene depth $d(\mathbf{x})$ of an image, and the atmospheric scattering coefficient $\beta$, $t$ can be calculated by $t(\mathbf{x})=\exp(-\beta d(\mathbf{x}))$. Based on this model, \cite{sun2015} is a traditional method, which simulates the haze with different visibility.

In addition, with the booming development of DNNs and generative adversarial networks (GAN), data-driven methods with different pipelines have been proposed to translate images from one domain to another, \ie, image-to-image translation, \eg, Pix2Pix \cite{pix2pix}, CycleGAN \cite{cyclegan}, BicycleGAN \cite{bicyclegan}, DualGAN \cite{dualgan}, MUNIT \cite{munit}, StarGAN \cite{stargan,starganv2}, U-GAT-IT \cite{ugatit}, \etc. GANs have achieved promising results on various image translation tasks and can be used in haze synthesis as well \cite{zhang2021}. Nevertheless, the haze synthesized by these methods can hardly be manipulated to launch adversarial attacks, as these methods are based on end-to-end networks and lack the ability to fine-tune the haze during an adversarial attack. Therefore, these methods are not suitable for our approach. Hence, these methods are not applicable for adversarial attack purposes.

As a result, we focus on Eq.~\eqref{eq:haze} that has been widely used to synthesize haze in an image in previous work, where $\mathbf{A}$ is usually considered as a constant across the whole image. However, when an uneven light source exists, this assumption becomes unsuitable, \eg, an image including influential sunlight. Also, we should take into account the inhomogeneous atmosphere, so the more general transmittance can be expressed as
\begin{align}
    t(\mathbf{x}) = \exp(-\beta(\mathbf{x})d(\mathbf{x})),
\end{align}
where $\beta(\mathbf{x})$ is the scattering coefficient of the atmosphere for location $\mathbf{x}$, and $d$ is the scene depth map. When the atmosphere is homogeneous, $\beta$ is a constant at all locations $\mathbf{x}$ in the image, and conversely $\beta$ can vary from location to location. We can acquire depth maps using specific capture devices, \eg, time of flight (ToF) sensors, or by estimating depth maps using efficient models, \eg, MiDaS \cite{midas}, monodepth \cite{monodepth}, monodepth2 \cite{monodepth2}, DPNet \cite{dpnet}, \etc. In adversarial attack scenarios, we take the second approach, which is more practical. Then, we denote $\hat{d} = D(\mathbf{I})$, where $D$ is a depth estimator. To sum up, the haze synthesis model can be denoted as
\begin{align}\label{eq:esthaze}
    \hat{\mathbf{H}}=\mathrm{haze}_D(\mathbf{I},\mathbf{A},\beta),
\end{align}
where $\hat{\mathbf{H}}$ is the hazy image estimated based on the predicted depth. Then, we introduce the adversarial haze attack.

\subsection{Adversarial Haze Attack}

Given a clean image $\mathbf{I}$ and a pre-trained DNN $\phi$, we aim to use Eq.~\eqref{eq:esthaze} to synthesize a hazy image that can mislead $\phi$ to predict an incorrect class. We can tune $\mathbf{A}$ and $\beta$ simultaneously to encourage this. Then, the proposed adversarial haze attack becomes a constrained optimization problem
\begin{align}\label{eq:advhaze}
\begin{aligned}
    \argmax_{\mathbf{A},\beta}\ &
    J(\phi(\mathrm{haze}_D(\mathbf{I},\mathbf{A},\beta)),y),
    \\
    \text{subject to}\ \forall\mathbf{x},\
    & {\|\mathbf{A}(\mathbf{x})-\mathbf{A}_0\|}_\infty<\epsilon_\mathbf{A},
    \\
    & {\|\beta(\mathbf{x})-\beta_0\|}_\infty<\epsilon_\beta,
\end{aligned}
\end{align}
where $\epsilon_\mathbf{A}$ and $\epsilon_\beta$ are the maximum perturbations in terms of the $l^\infty$ norm balls and $\mathbf{A}_0$ and $\beta_0$ are the initialized values for $\mathbf{A}$ and $\beta$, respectively; $y$ denotes the true label of the clean image $\mathbf{I}$, and $J$ is the classification loss function for $\phi$. In this paper, we employ the cross-entropy loss ($\mathcal{L}_\mathrm{CE}$, widely used in the image classification task) as the loss function, \ie,
\begin{align}
    J(\hat{y},y)=\mathcal{L}_{\mathrm{CE}}=-\sum_{i=1}^{N}y_i\log\hat{y}_i,
\end{align}
where $N$ denotes the number of classes, $\hat{y}_i$ is the predicted score for class $i$, and $y$ is the one-hot encoding of the true label.

Nevertheless, the adversarial examples generated based on Eq.~\eqref{eq:advhaze} are usually noisy, because pixel-wise tuning is allowed to $\mathbf{A}$ and $\beta$ without spatial constraints. To overcome this issue, we propose two variants of the adversarial haze attack based on different assumptions.

\subsubsection{Homogeneous Adversarial Haze (HAdvHaze).} We assume that the global atmospheric light is even, and the atmosphere is homogeneous. For each location $\mathbf{x}$, we have constant $\mathbf{A}(\mathbf{x})=\tilde{\mathbf{A}}$ and $\beta(\mathbf{x})=\tilde{\beta}$. Then, Eq.~\eqref{eq:advhaze} becomes
\begin{align}\label{eq:hadvhaze}
\begin{aligned}
    \argmax_{\tilde{\mathbf{A}},\tilde{\beta}}\ &
    J(\phi(\mathrm{haze}_D(\mathbf{I},\tilde{\mathbf{A}},\tilde{\beta})),y),
    \\
    \text{subject to}\ \forall\mathbf{x},\ 
    &{\|\tilde{\mathbf{A}}-\mathbf{A}_0\|}_\infty<\epsilon_\mathbf{A},
    \\
    &{\|\tilde{\beta}-\beta_0\|}_\infty<\epsilon_\beta.
\end{aligned}
\end{align}

\begin{figure}[t]
\centering
\includegraphics[width=0.9\textwidth]{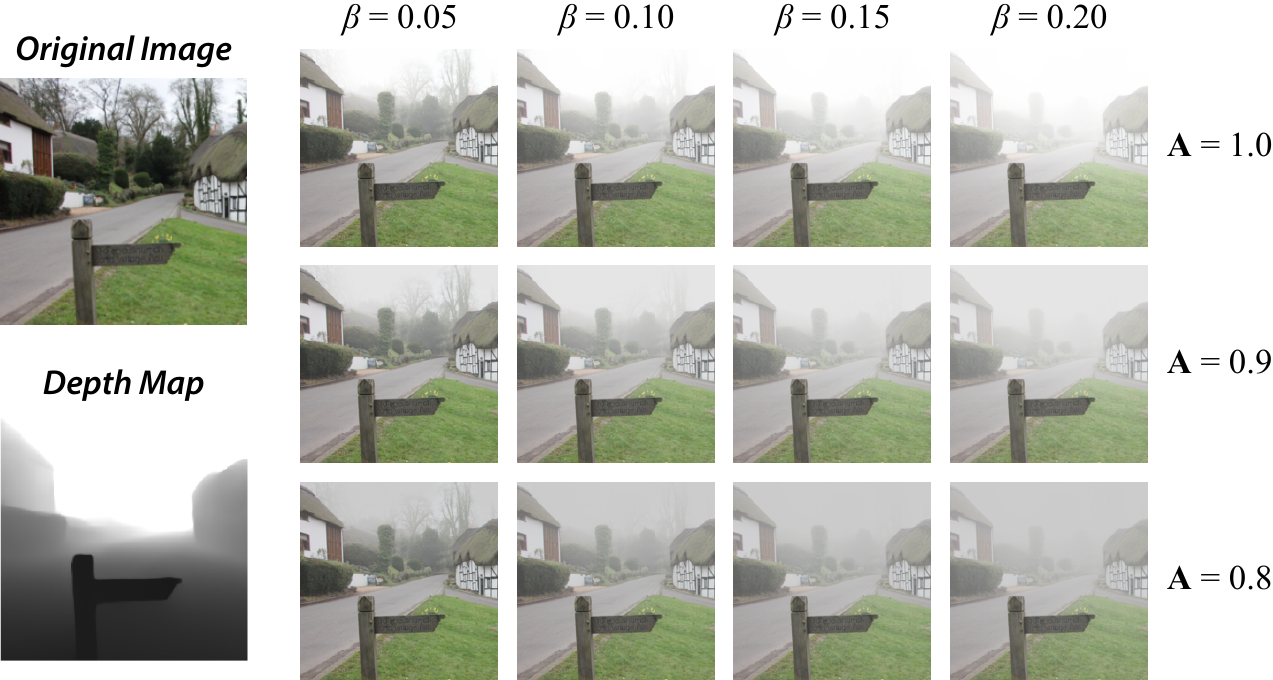}
\caption{Synthesized hazy images, assuming the global atmospheric light is even, and the atmosphere is homogeneous. We take values for $\mathbf{A}$ in $\{0.8,0.9,1.0\}$ and $\beta$ in $\{0.05,0.10,0.15,0.20\}$. The haze is uniform in image areas of equal depth, and visibility decreases with increasing scene depth.}
\label{fig:homo}
\end{figure}

However, this variant holds an overly strict assumption, \ie, that the global atmospheric light is even, and the atmosphere is homogeneous. We visualize the images synthesized under this setting in Figure~\ref{fig:homo}. Intuitively, it's difficult for the adversarial samples with only two global parameters tuned to fool classifiers. This is demonstrated by the quantitative experimental results shown in Section~\ref{sec:expperf} and \ref{sec:exptran}.

\subsubsection{Inhomogeneous Adversarial Haze (IAdvHaze).} In a wild scene, the assumption for Eq.~\eqref{eq:hadvhaze} is easily broken. That is to say, the light sources in the scene are uneven and the atmosphere is inhomogeneous, which allows us to locally tune $\mathbf{A}$ and $\beta$, respectively. However, for a real captured image, these two parameters are locally smooth. To this end, we apply low-pass filters $f_\mathbf{A}$ and $f_\beta$ to pixel-wise tune $\mathbf{A}'$ and $\beta'$, \ie, $\mathbf{A}=\mathbf{A}'\ast f_\mathbf{A}$ and $\beta=\beta'\ast f_\mathbf{\beta}$, where $\ast$ stands for the convolution operation. Then, Eq.~\eqref{eq:advhaze} becomes
\begin{align}\label{eq:iadvhaze}
\begin{aligned}
    \argmax_{\mathbf{A}',\beta'}\ &
    J(\phi(\mathrm{haze}_D(\mathbf{I},\mathbf{A}'\ast f_\mathbf{A},\beta'\ast f_\mathbf{\beta}),y),
    \\
    \text{subject to}\ \forall\mathbf{x},\ 
    &{\|(\mathbf{A}'\ast f_\mathbf{A})(\mathbf{x})-\mathbf{A}_0\|}_\infty<\epsilon_\mathbf{A},
    \\
    &{\|(\beta'\ast f_\mathbf{\beta})(\mathbf{x})-\beta_0\|}_\infty<\epsilon_\beta.
\end{aligned}
\end{align}

\begin{figure}[t]
\centering
\includegraphics[width=0.8\textwidth]{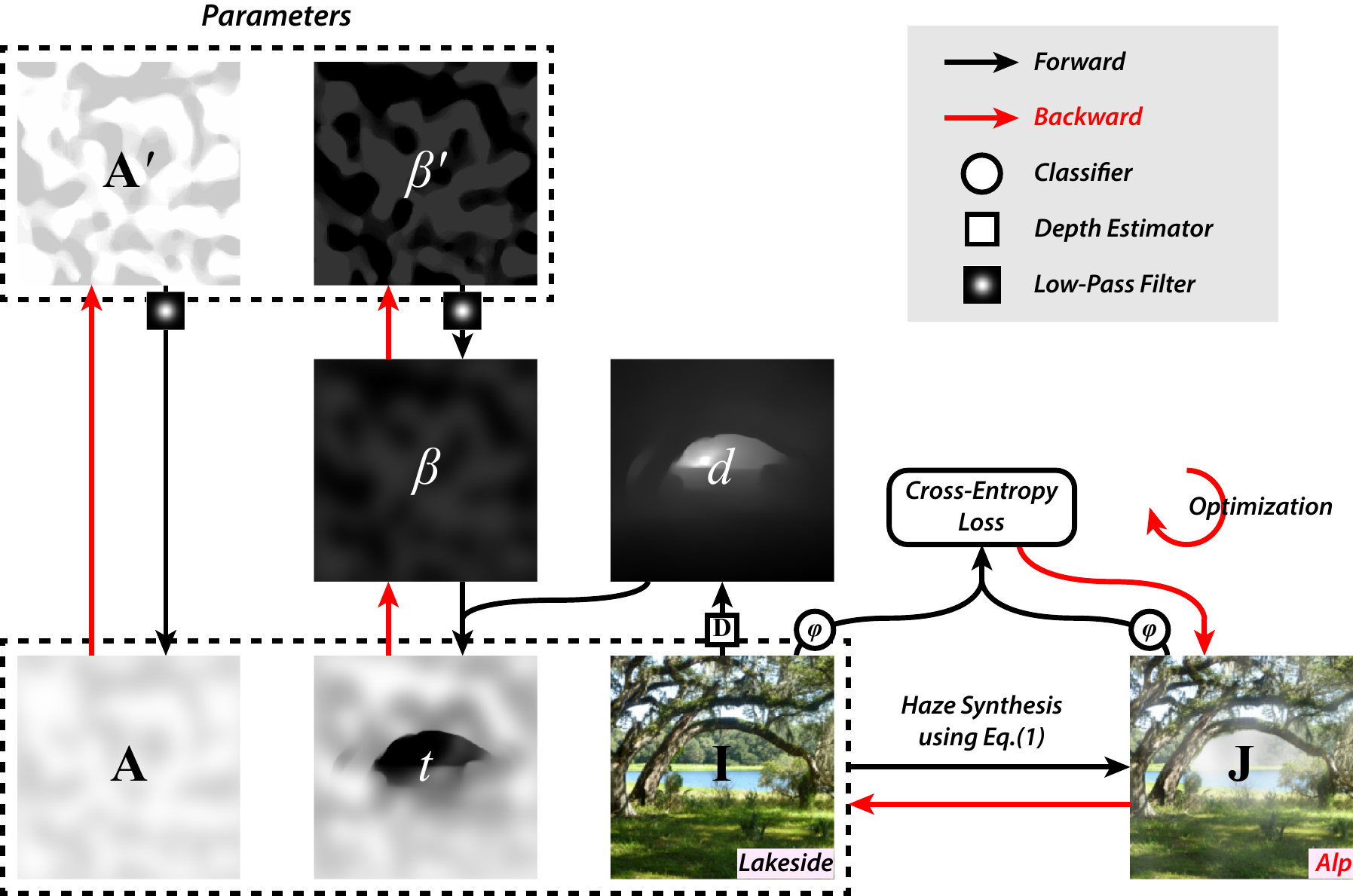}
\caption{An illustration of the inhomogeneous adversarial haze's pipeline. The scene depth of $\mathbf{I}$ is obtained by a depth estimator. We use a backpropagation algorithm to calculate the gradient of the parameters, \ie, $\mathbf{A}'$ and $\beta'$, against the cross-entropy loss. The parameters are optimized through a gradient descent method and succeed to fool a classifier.}\label{fig:pipeline}
\end{figure}

To further explain how the inhomogeneous adversarial haze works, we also show an illustration in Figure~\ref{fig:pipeline}. In this example, our adversarial attack succeeds in getting a classifier to recognize ``lakeside'' as ``alp''.

\subsubsection{Implementation Details.}

We can simply use a sign gradient descent method to optimize Eq.~\eqref{eq:hadvhaze} and Eq.~\eqref{eq:iadvhaze} in this paper. We name these two variants HAdvHaze and IAdvHaze in Section~\ref{sec:exp} respectively. More specifically, we employ MI-FGSM \cite{mifgsm} with iteration number $n=10$, momentum $\mu=1.0$, $\mathbf{A}_0=0.9$, $\beta_0=0.1$, and step sizes $\alpha_\mathbf{A}=\alpha_\beta=0.01$, respectively. Note that, we treat $\mathbf{A}(\mathbf{x})$ as a scalar parameter, \ie, each channels of $\mathbf{A}(\mathbf{x})$ is the same value for a specific location $\mathbf{x}$. This avoids colorful haze being synthesized, which is usually strange. We employ MiDaS \cite{midas} as $D$ to estimate scene depth map $d(\mathbf{x})$ and use two Gaussian filters as $f_\mathbf{A}$ and $f_\beta$ for the IAdvHaze variant.

\section{Experiments}\label{sec:exp}

In this section, we perform several experiments to demonstrate the effectiveness of the proposed adversarial attack. First, we report our experimental setups in Section~\ref{sec:expset}. Then, we compare our method with baselines on performance, transferability, and image quality in both quantitative and visualized results in Section~\ref{sec:expperf} and \ref{sec:exptran}. We also show correlation matrices among attacks, to further discuss the relationship between noise-based and non-noise-based attacks in Section~\ref{sec:exptran}.

\subsection{Experimental Setups}\label{sec:expset}

\subsubsection{Datasets.}

We perform experiments on ImageNet \cite{imagenet} and NIPS~2017 \cite{nips2017} to evaluate baselines and our attack for comparison. ImageNet is a dataset for developing and evaluating object detection and image classification methods at a large scale. We use the validation split of its ILSVRC~2012 subset, which contains 50,000 images of 1,000 categories. NIPS~2017 (Adversarial Learning Development Set) is the development images used in the NIPS~2017 Adversarial Learning challenges, which aims to accelerate research on adversarial examples. NIPS~2017 contains 1,000 labeled images.

\subsubsection{Target Models.}

We employ four popular deep neural network models for image classification, \ie, DenseNet-161 \cite{densenet}, ResNet-50 \cite{resnet}, Inception-V3 \cite{inceptionv3}, MobileNet-V2 \cite{mobilenetv2}, as the target models. We will also transfer the generated adversarial examples to the other models in addition to the target model alone, to further evaluate the transferability of adversarial attacks.

\subsubsection{Baseline Attacks.}

We collect four additive noise attack methods, \ie, FGSM \cite{fgsm}, I-FGSM \cite{ifgsm}, MI-FGSM \cite{mifgsm}, TI-MI-FGSM \cite{ti}, and one colorization attack method, \ie, cAdv \cite{cadv}, as the baselines. We map the image pixel values of each channel to $[0,1]$. For noise attacks, we set their hyper-parameters as maximum perturbation $\epsilon=10/255$, and iteration number $n=10$ for the attacks that are based on iteration.

\subsubsection{Metrics.}

We follow the commonly used metric, \ie, success rate $\mathcal{S}$, to evaluate the performance of baselines and the proposed attack. It is calculated by dividing the misclassified images' number by the total number. Also, we adopt a blind image quality assessor, \ie, NIQE \cite{niqe}, for measuring the image quality of the generated adversarial examples. A lower NIQE score indicates a better quality of an image. Under these metrics, we should balance both the success rate and the image quality.

\subsection{Comparison on Performance}\label{sec:expperf}

\begin{table}[t]
\caption{Performance and image quality results of white-box adversarial attacks. For different target models and datasets, we report white-box attack success rates and NIQE scores of the generated adversarial examples in the Succ. and NIQE columns, respectively. We highlight the proposed attacks in red and bold the NIQE scores that are much larger than 5.}\label{tab:atk}
\centering
\adjustbox{width=0.7\textwidth}{
\begin{tabular}{c|c|cc|cc|cc|cc}
\toprule
\multirow{2}{*}{}                 &
Targets                           &
\multicolumn{2}{c|}{DenseNet-161} &
\multicolumn{2}{c|}{ResNet-50}    &
\multicolumn{2}{c|}{Inception-V3} &
\multicolumn{2}{c}{MobileNet-V2}  \\
\cmidrule{2-10}
&
Metrics &
Succ.   &
NIQE    &
Succ.   &
NIQE    &
Succ.   &
NIQE    &
Succ.   &
NIQE    \\
\midrule
\multirow{7}{*}{\rotatebox[origin=c]{90}{ImageNet}}
&
FGSM            &
86.64\%         &
\lniqe{21.59}   &
85.53\%         &
\lniqe{25.89}   &
79.18\%         &
\lniqe{16.26}   &
94.26\%         &
\lniqe{34.72}   \\
&
I-FGSM          &
99.72\%         &
5.07            &
99.97\%         &
5.19            &
98.44\%         &
4.80            &
100.00\%        &
5.34            \\
&
MI-FGSM         &
99.71\%         &
\lniqe{8.13}    &
99.98\%         &
\lniqe{8.84}    &
98.36\%         &
\lniqe{7.20}    &
100.00\%        &
\lniqe{10.27}   \\
&
TI-MI-FGSM      &
99.47\%         &
5.04            &
99.79\%         &
5.02            &
97.55\%         &
4.94            &
99.99\%         &
5.02            \\
&
cAdv            &
97.77\%         &
4.97            &
97.17\%         &
4.96            &
92.21\%         &
4.97            &
98.72\%         &
4.96            \\
&
\ours
HAdvHaze        &
\ours
2.88\%          &
\ours
5.01            &
\ours
5.37\%          &
\ours
5.00            &
\ours
8.15\%          &
\ours
5.00            &
\ours
9.07\%          &
\ours
5.01            \\
&
\ours
IAdvHaze        &
\ours
99.22\%         &
\ours
5.68            &
\ours
99.48\%         &
\ours
5.69            &
\ours
96.40\%         &
\ours
5.66            &
\ours
99.95\%         &
\ours
5.65            \\
\midrule
\multirow{7}{*}{\rotatebox[origin=c]{90}{NIPS~2017}}
&
FGSM            &
90.16\%         &
\lniqe{23.60}   &
90.01\%         &
\lniqe{27.95}   &
88.02\%         &
\lniqe{17.48}   &
94.15\%         &
\lniqe{39.70}   \\
&
I-FGSM          &
99.86\%         &
5.01            &
100.00\%        &
5.19            &
99.86\%         &
4.80            &
100.00\%        &
5.35            \\
&
MI-FGSM         &
99.86\%         &
\lniqe{8.54}    &
100.00\%        &
\lniqe{9.28}    &
99.86\%         &
\lniqe{7.58}    &
100.00\%        &
\lniqe{11.02}   \\
&
TI-MI-FGSM      &
99.86\%         &
5.14            &
99.86\%         &
5.09            &
99.57\%         &
4.99            &
100.00\%        &
5.13            \\
&
cAdv            &
98.72\%         &
5.09            &
98.00\%         &
5.06            &
96.43\%         &
5.08            &
98.86\%         &
5.05            \\
&
\ours
HAdvHaze        &
\ours
6.28\%          &
\ours
5.22            &
\ours
11.84\%         &
\ours
5.19            &
\ours
15.26\%         &
\ours
5.22            &
\ours
17.40\%         &
\ours
5.22            \\
&
\ours
IAdvHaze        &
\ours
99.86\%         &
\ours
5.82            &
\ours
99.57\%         &
\ours
5.79            &
\ours
99.43\%         &
\ours
5.78            &
\ours
99.86\%         &
\ours
5.76            \\
\bottomrule
\end{tabular}
}
\end{table}

We launch all baseline attacks and our proposed attacks on two datasets and report the white-box success rates and the NIQE scores in Table~\ref{tab:atk}. We have two findings: (1) All the attacks achieve nearly 100\% success rates to fool the target models, except HAdvHaze. It is difficult to fool a classifier for its only two scalar parameters to be tuned during the attack iterations. (2) FGSM and MI-FGSM hold the two largest NIQE scores at most time, which means that its generated adversarial examples are of the worst image quality. Nevertheless, the proposed IAdvHaze with much lower NIQE scores achieves competitive performance on the ImageNet dataset and outperforms them on the NIPS~2017 dataset, as shown in Section~\ref{sec:exptran}. The other adversarial attacks have NIQE scores around 5. (3) The NIQE scores cannot reflect the image quality perfectly, for example, the TI-MI-FGSM introduces regional noise patterns, and in fact, this perturbation is easily perceived, as we discuss in the following paragraph.

\begin{figure}[t!]
\centering
\includegraphics[width=\textwidth]{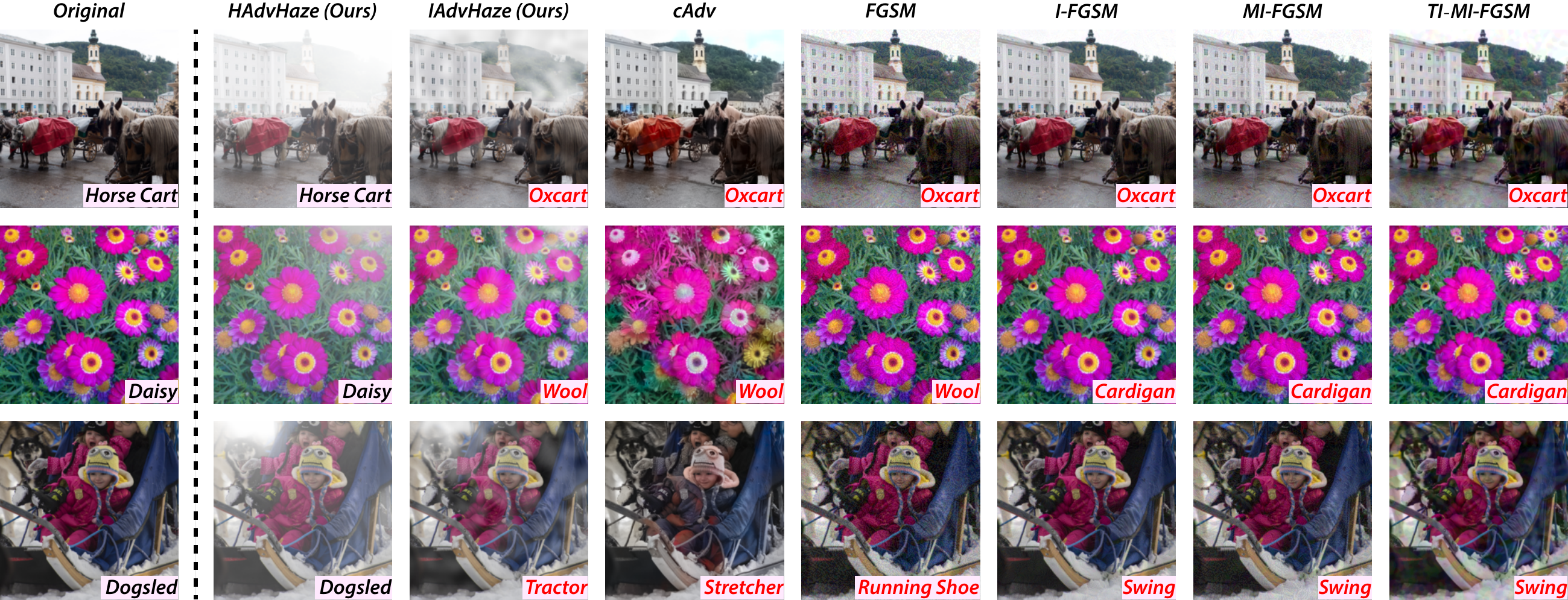}
\caption{Visualized cases, \ie, adversarial examples generated by different attacks. We show the classes predicted by the target classifier in the lower right corner of each example, respectively. We use \textcolor{red}{red} to highlight the incorrectly predicted classes.
}\label{fig:vis}
\end{figure}

To compare the image quality of adversarial examples directly, we also show a couple of visualization cases in Figure~\ref{fig:vis}. We can visually find that our generated haze phenomenon is the most natural perturbation. In contrast, the noise-based adversarial examples contain strange noise patterns unavoidably; the colorization attack is better than the noise-based attack, but it can also accidentally turn several semantic regions into unnatural colors, \eg, the leaves in the second visualization case are modified to an unnatural red color.

\subsection{Comparison on Transferability}\label{sec:exptran}

\begin{table}[t]
\caption{Transferability results of adversarial attacks. We transfer all launched attacks to the other models instead of their targets, \eg, transfer attacks targeting DenseNet-161 (DN161) to ResNet-50 (Res50), Inception-V3 (IncV3), and MobileNet-V2 (MobV2) on the ImageNet dataset and the NIPS~2017 dataset. We report the transferablity performance in success rates and highlight the top three attacks in \rf{red}, \rs{green} and \rt{blue} respectively.}\label{tab:transfer_atk}
\centering
\adjustbox{width=\textwidth}{
\begin{tabular}{c|c|ccc|ccc|ccc|ccc}
\toprule
\multirow{2}{*}{}                 &
Targets                           &
\multicolumn{3}{c|}{DenseNet-161} &
\multicolumn{3}{c|}{ResNet-50}    &
\multicolumn{3}{c|}{Inception-V3} &
\multicolumn{3}{c}{MobileNet-V2}  \\
\cmidrule{2-14}
&
Transfer to &
Res50 &
IncV3 &
MobV2 &
DN161 &
IncV3 &
MobV2 &
DN161 &
Res50 &
MobV2 &
DN161 &
Res50 &
IncV3 \\
\midrule
\multirow{7}{*}{\rotatebox{90}{ImageNet}}
&
FGSM            &
41.84\%         &
31.94\%         &
\rt{43.97\%}    &
34.43\%         &
29.15\%         &
42.07\%         &
\rs{21.16\%}    &
\rs{25.40\%}    &
\rf{34.79\%}    &
23.54\%         &
31.34\%         &
26.81\%         \\
&
I-FGSM          &
\rt{50.65\%}    &
19.96\%         &
37.29\%         &
35.70\%         &
14.97\%         &
35.36\%         &
8.16\%          &
9.68\%          &
14.21\%         &
15.08\%         &
22.06\%         &
12.46\%         \\
&
MI-FGSM         &
\rf{71.12\%}    &
\rs{43.04\%}    &
\rf{59.91\%}    &
\rf{65.17\%}    &
\rs{37.15\%}    &
\rf{60.37\%}    &
\rf{22.42\%}    &
\rf{25.48\%}    &
\rs{32.72\%}    &
\rf{36.82\%}    &
\rf{47.16\%}    &
\rt{32.39\%}    \\
&
TI-MI-FGSM      &
\rs{50.69\%}    &
\rt{34.61\%}    &
\rs{45.62\%}    &
\rs{47.17\%}    &
\rt{31.44\%}    &
\rs{47.23\%}    &
16.63\%         &
18.01\%         &
24.11\%         &
\rs{35.84\%}    &
\rs{41.92\%}    &
\rs{32.67\%}    \\
&
cAdv            &
26.43\%         &
21.57\%         &
25.97\%         &
25.17\%         &
20.70\%         &
29.52\%         &
10.20\%         &
11.51\%         &
14.82\%         &
21.34\%         &
25.85\%         &
20.98\%         \\
&
HAdvHaze        &
4.00\%          &
5.91\%          &
6.51\%          &
2.37\%          &
6.41\%          &
7.40\%          &
2.04\%          &
3.58\%          &
6.09\%          &
2.34\%          &
4.37\%          &
6.73\%          \\
&
IAdvHaze        &
45.15\%         &
\rf{44.04\%}    &
42.04\%         &
\rt{44.49\%}    &
\rf{43.47\%}    &
\rt{46.24\%}    &
\rt{19.63\%}    &
\rt{21.79\%}    &
\rt{27.23\%}    &
\rt{32.81\%}    &
\rt{39.32\%}    &
\rf{41.82\%}    \\
\midrule
\multirow{7}{*}{\rotatebox{90}{NIPS~2017}}
&
FGSM            &
48.22\%         &
\rt{39.66\%}    &
47.65\%         &
39.80\%         &
36.80\%         &
44.37\%         &
\rs{26.53\%}    &
\rf{31.10\%}    &
\rs{39.37\%}    &
27.67\%         &
35.66\%         &
31.67\%         \\
&
I-FGSM          &
\rt{56.49\%}    &
25.11\%         &
37.80\%         &
39.80\%         &
19.26\%         &
38.09\%         &
8.13\%          &
9.27\%          &
16.12\%         &
14.84\%         &
25.11\%         &
17.26\%         \\
&
MI-FGSM         &
\rf{77.60\%}    &
\rs{49.79\%}    &
\rf{63.34\%}    &
\rf{72.47\%}    &
\rs{45.36\%}    &
\rf{63.48\%}    &
\rt{23.40\%}    &
\rs{28.67\%}    &
\rt{36.09\%}    &
\rs{40.23\%}    &
\rs{51.36\%}    &
\rt{36.23\%}    \\
&
TI-MI-FGSM      &
55.78\%         &
39.51\%         &
\rt{48.79\%}    &
\rt{52.92\%}    &
\rt{40.80\%}    &
\rt{52.35\%}    &
17.26\%         &
19.40\%         &
31.10\%         &
\rt{39.23\%}    &
\rt{46.08\%}    &
\rs{41.37\%}    \\
&
cAdv            &
26.96\%         &
28.10\%         &
30.24\%         &
28.39\%         &
25.11\%         &
32.24\%         &
11.13\%         &
12.27\%         &
15.83\%         &
24.54\%         &
28.67\%         &
28.53\%         \\
&
HAdvHaze        &
9.13\%          &
12.84\%         &
12.41\%         &
4.71\%          &
13.84\%         &
15.26\%         &
4.56\%          &
6.85\%          &
11.55\%         &
5.42\%          &
9.56\%          &
13.98\%         \\
&
IAdvHaze        &
\rs{60.20\%}    &
\rf{60.77\%}    &
\rs{56.21\%}    &
\rs{59.34\%}    &
\rf{57.06\%}    &
\rs{58.35\%}    &
\rf{29.39\%}    &
\rt{28.53\%}    &
\rf{39.80\%}    &
\rf{46.22\%}    &
\rf{55.35\%}    &
\rf{58.49\%}    \\
\bottomrule
\end{tabular}
}
\end{table}

We compare the two variants of the proposed attack with baselines on transferability and show the performance results in Table~\ref{tab:transfer_atk}. For the ImageNet dataset, the proposed IAdvHaze achieves competitive transfer performance, compared with noise-based adversarial attacks, \eg, MI-FGSM and TI-MI-FGSM. Note that, FGSM and MI-FGSM hold worse scores in NIQE than the proposed attack and the adversarial examples generated by TI-MI-FGSM contain perceptible regional noise patterns. For the NIPS~2017 dataset, the proposed IAdvHaze outperforms the baselines most of the time, especially when transferred to the Inception-V3 classifier.

\begin{figure}[t!]
\centering
\includegraphics[width=\textwidth]{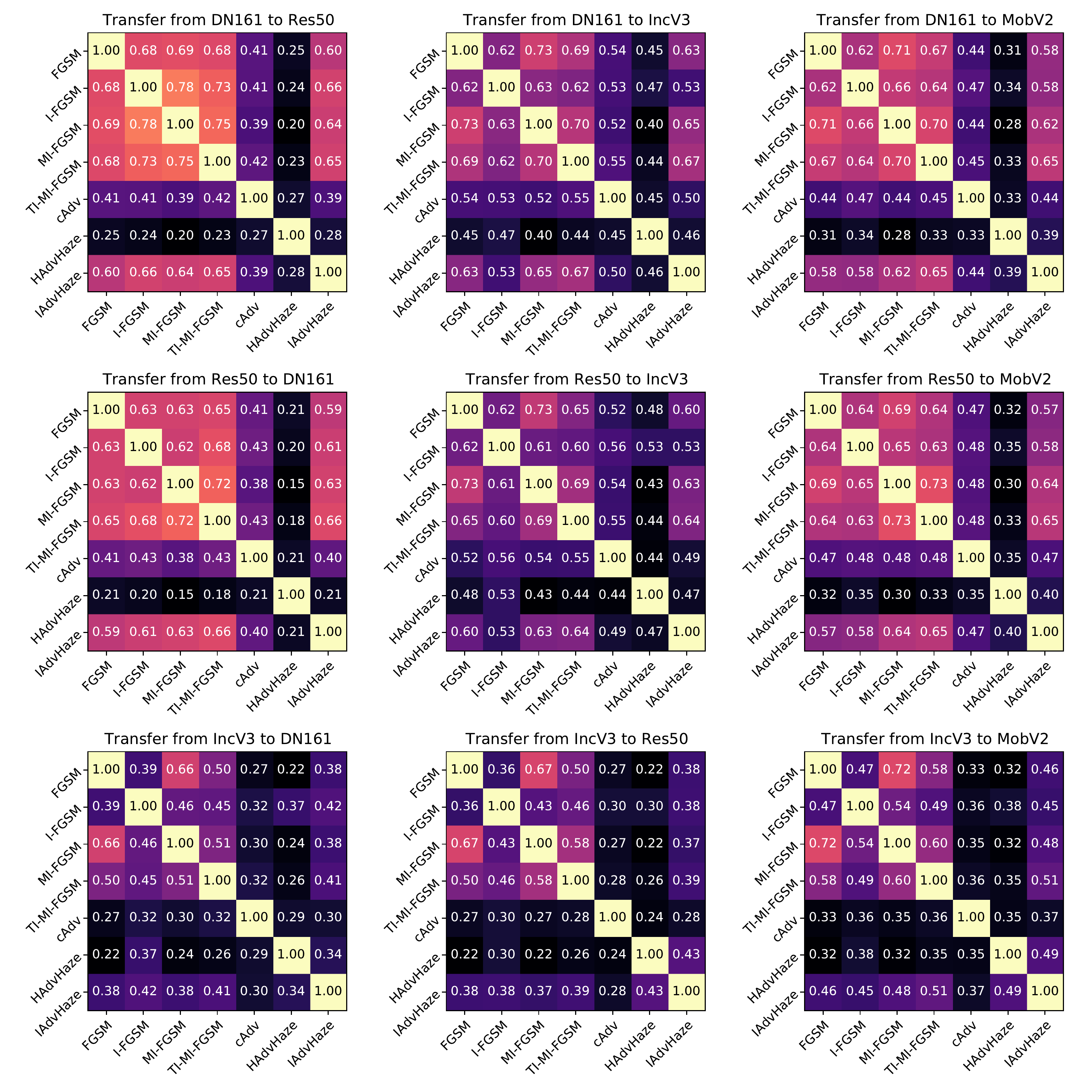}
\caption{Correlation matrices of transfer attacks on the NIPS~2017 dataset. We transfer each attack to its non-targeted models and calculate IoU ratios of success cases as the correlation between every two attacks. We aim to identify the relationship between noise-based and non-noise based attacks in a visual way.}\label{fig:iou}
\end{figure}

To quantitatively reflect the relationship among attacks, we show the correlation matrices in Figure~\ref{fig:iou}, in which the intersection over union (IoU) ratio of success cases between every two attacks describes their correlation. We find that non-noise-based attacks, \ie, cAdv, HAdvHaze, and IAdvHaze, usually hold less IoU ratio to noise-based attacks, \ie, FGSM, I-FGSM, MI-FGSM, and TI-MI-FGSM, especially when transferred from Inception-V3. To some extent, this demonstrates that non-noise-based adversarial attacks' success cases cover a different range of images from noise-based attacks. We try to explain this phenomenon intuitively: noise-based attacks carefully add noise to images using small step sizes, which tend to trap images in local minima. In contrast, non-noise-based attacks apply perturbations in completely different ways, \eg, haze synthesis and colorization, and usually have larger equivalent step sizes in the sense of noise attacks. Inspired by this, we may jointly apply various perturbations to achieve a more effective attack against DNNs in future work.

\section{Conclusion}

We propose a novel adversarial attack, which can fool varied classifiers to predict a wrong category. The proposed method is different from common noise-based attacks and instead synthesizes uneven natural haze to generate adversarial examples. We conduct comprehensive experiments and make quantitative comparisons to verify the efficiency of our method. Inspired by the correlation matrices, we present an idea to jointly apply different perturbations to improve the success rate of the attack. In future work, we will bring the idea of joint perturbations to the field and go further in developing non-noise-based adversarial attack methods. We hope that these novel non-noise-based adversarial attacks can be used to evaluate and improve the security and robustness of DNNs.

\bibliographystyle{splncs04}
\bibliography{ref.bib}
\end{document}